\documentclass[conference]{IEEEtran}

\ifCLASSOPTIONcompsoc
  \usepackage[nocompress]{cite}
\else
  \usepackage{cite}
\fi

\ifCLASSINFOpdf
\usepackage[pdftex]{graphicx}
\else
\fi

\usepackage[float]{}

\usepackage{algorithm}% http://ctan.org/pkg/algorithms
\usepackage{algpseudocode}% http://ctan.org/pkg/algorithmicx
\usepackage{textcomp}
\usepackage[english]{babel}
\usepackage[utf8x]{inputenc}
\usepackage{amsmath}
\usepackage{amssymb}
\usepackage{graphicx}
\usepackage[export]{adjustbox}
\usepackage{array,multirow}
\usepackage[table]{xcolor}
\usepackage{listings}
\usepackage{cleveref}
\usepackage{url}
\usepackage{arydshln}
\usepackage{todonotes}
\graphicspath{ {figures/} }
\makeatletter
\DeclareTextCommandDefault{\textleftarrow}{\mbox{$\m@th\leftarrow$}}
\makeatother

\begin{document}

\title{Rolling Horizon NEAT\\ for General Video Game Playing}

\IEEEoverridecommandlockouts
\IEEEpubid{\begin{minipage}{\textwidth}\ \\[12pt]
978-1-7281-4533-4/20/\$31.00 \copyright 2020 IEEE
\end{minipage}}

\author{
\IEEEauthorblockN{Diego Perez-Liebana}
\IEEEauthorblockA{Queen Mary University of London\\
Game AI Group\\
London, UK\\
diego.perez@qmul.ac.uk}
\and
\IEEEauthorblockN{Muhammad Sajid Alam}
\IEEEauthorblockA{Queen Mary University of London\\
Game AI Group\\
London, UK\\
m.s.alam@se16.qmul.ac.uk}
\and
\IEEEauthorblockN{Raluca D. Gaina}
\IEEEauthorblockA{Queen Mary University of London\\
Game AI Group\\
London, UK\\
r.d.gaina@qmul.ac.uk}}

\IEEEtitleabstractindextext{%
\begin{abstract}
This paper presents a new Statistical Forward Planning (SFP) method, Rolling Horizon NeuroEvolution of Augmenting Topologies (rhNEAT). Unlike traditional Rolling Horizon Evolution, where an evolutionary algorithm is in charge of evolving a sequence of actions, rhNEAT evolves weights and connections of a neural network in real-time, planning several steps ahead before returning an action to execute in the game. Different versions of the algorithm are explored in a collection of 20 GVGAI games, and compared with other SFP methods and state of the art results. Although results are overall not better than other SFP methods, the nature of rhNEAT to adapt to changing game features has allowed to establish new state of the art records in games that other methods have traditionally struggled with. The algorithm proposed here is general and introduces a new way of representing information within rolling horizon evolution techniques.
\end{abstract}
}

\maketitle

\IEEEdisplaynontitleabstractindextext

%%%%%%%%%%%%%%%%%%%%%%%%%%%%%%%%%%%%%%%
\section{Introduction} \label{sec:intro}

Research in General Video Game Playing (GVGP) has become very popular in the last years, with the proliferation of frameworks and studies around the Atari Learning Environment (ALE;~\cite{genesereth2005general}), the General Video Game AI (GVGAI;~\cite{perez20152014}) framework and many other benchmarks. Either from a learning or from a planning point of view, an important body of research has focused on generalization of game-playing agents for multiple games. The GVGAI framework has featured in multiple recent studies on general game playing. While its main alternative is ALE, GVGAI provides a higher variety of scenarios and levels per game: via the Video Game Description Language (VGDL), games are easily extensible and a potentially infinite number of games and levels can be created.

Among the most popular approaches used in GVGAI for real-time planning, Statistical Forward Planning (SFP) methods have a distinguished position~\cite{perez2019general}. SFP techniques include Monte Carlo Tree Search (MCTS;~\cite{browne2014MCTSsurvey}) and Rolling Horizon Evolutionary Algorithms (RHEA;~\cite{perez2013rolling}), and multiple variants of these approaches can be found in the GVGAI literature. Most of these methods do not use game features, some exceptions being Knowledge-Based MCTS~\cite{perez2014knowledge} and one of the competition winners YOLOBOT~\cite{joppen2017informed}. In both cases, game features are used to either prune branches, bias rollouts or evaluate game states, with generally good results~\cite{gvgaibook2019}.

The objective of this paper is to dive deeper into this type of mixed approaches, by introducing a new SFP method. We present \textit{rhNEAT}, an algorithm that takes concepts from the existing NeuroEvolution of Augmenting Topologies (NEAT) and RHEA. We analyze different variants of the algorithm and compare its performance with state of the art results. A second contribution of this paper is to start opening a line of research that explores alternative representations for Rolling Horizon Evolutionary Algorithms: while traditionally RHEA evolves a sequence of actions, rhNEAT evolves neural network weights and configurations, which in turn generate action sequences used to evaluate individual fitness.

Background concepts are described in Section~\ref{sec:back}, with a special focus on NEAT in Section~\ref{sec:neat}. The proposed rhNEAT approach is detailed in Section~\ref{sec:methods}. Experiments and results are described in Sections~\ref{sec:exp} and~\ref{sec:results}, before concluding the paper in Section~\ref{sec:end}.

%%%%%%%%%%%%%%%%%%%%%%%%%%%%%%%%%%%%%%%
\section{Background} \label{sec:back}

\subsection{GVGAI}

The General Video Game Artificial Intelligence (GVGAI) framework and competition~\cite{perez20152014} focus on the application of AI techniques for procedural content generation and game playing in multiple games. Rather than specialising in a single game, GVGAI fosters AI research in a collection of 2D arcade games minimizing the use of domain knowledge. The framework currently has more than $180$ single and two-player games.

In the game planning tracks of GVGAI, agents can interact with the game in real time to provide actions to execute at every frame. During this thinking time, agents can access a reduced observation of the environment, including game score, game state (win, loss or ongoing), current time step and player (or avatar) status (orientation, position resources and health points). The agent is also provided with the list of available actions and observations of other sprites. These sprites are grouped into certain categories: non-player characters (NPC), immovable/movable sprites, resources that can be collected, portal sprites (that teleport, spawn or destroy other game elements) and other sprites created by the avatar.

The agent is also provided with a Forward Model (FM), which can reach a future state $s_{t+1}$ upon receiving a previous game state $s_t$ and an action $a$. The FM allows the implementation of planning approaches such as MCTS and RHEA, which form the base of most methods researchers have tested in GVGAI. For more details on GVGAI, its different tracks, competition editions and approaches, the reader is referred to the recent GVGAI survey~\cite{perez2019general}.

\subsection{Statistical Forward Planning Methods}

Statistical Forward Planning (SFP) refers to a family of methods that use a Forward Model (FM) to simulate the effect of actions executed in copies of the current game state. This term is coined to gather under the same umbrella two family of methods that use FM sampling for decision making: tree search (predominantly MCTS) and RHEA variants.

\subsubsection{Monte Carlo Tree Search} MCTS is a well known game playing algorithm that builds an asymmetric tree in memory, balancing exploration of different actions and exploitation of the most promising ones. MCTS has been widely used in multiple domains~\cite{browne2014MCTSsurvey}, and at present stands as one of the top approaches for GVGAI~\cite{gvgaibook2019}. The algorithm iterates through the following four steps until a decision budget expires: \textit{tree selection} (where the tree is navigated balancing exploration and exploitation), \textit{expansion} (a new node is added to the tree), \textit{simulation} or \textit{rollout} (a Monte Carlo simulation that chooses actions at random until a determined depth is reached or the game is over) and \textit{backpropagation} (which updates the average reward of each traversed node with a value assigned to the state found at the end of the rollout). 

\subsubsection{Rolling Horizon Evolutionary Algorithms} RHEA~\cite{perez2013rolling} evolves individuals represented by a sequence of actions at every game tick. Using the FM, each action in the sequence is iteratively used to advance the game state $s_t$ to $s_{t+1}$ until the last action of the individual takes the game to a state that is evaluated. Once the decision budget is exhausted, the first action of the best individual is played in the game for that frame. RHEA has been widely used in General Video Game Playing~\cite{gaina2017analysis}\cite{gaina2017enhancements}\cite{gaina2019tackling} and in some cases it has been proven superior to MCTS, winning several editions of the GVGAI competition~\cite{perez2019general}.

\subsection{Neuroevolution in Games} \label{sec:neat}

% NeuroEvolution (NE) is an old field of research that arose in the late 1980s~\cite{pfeifer1989dynamics} as an alternative to backpropagation, when this mechanism was not deemed successful for Neural Network (NN) learning. In most cases, NE consisted of the evolution via genetic algorithms or evolutionary strategies of the weights on a NN to tackle a learning task. Although some earlier works concerned the evolution of NN topologies~\cite{miller1989designing}, Stanley and Miikkulainen proposed one of the most popular NE systems: NeuroEvolution of Augmented Topologies (NEAT)~\cite{stanley2002evolving}. In NEAT, en evolutionary algorithm evolves both the weights and the topology of a NN using a direct encoding, where a binary vector determines if a connection between two neurons exists or not.

NeuroEvolution (NE), the evolution via genetic algorithms or evolutionary strategies of the weights and connections of a Neural Network (NN), has been extensively used in games~\cite{baldominos2019automated}. Stanley and Miikkulainen proposed one of the most popular NE systems: NeuroEvolution of Augmented Topologies (NEAT)~\cite{stanley2002evolving}. NEAT uses a direct encoding, where a binary vector determines if a connection between two neurons exists or not, to evolve both the weights and the topology of a NN. Later, Stanley et al.~\cite{stanley2005real} evolved game playing agents in real-time for the game Neuroevolving Robotic Operatives (NERO), where the player trains a group of combat robots to play against other players' teams. In~\cite{gauci2010autonomous}, Gauci et al. propose HyperNEAT, an algorithm that evolves topology and parameters of a NN to play checkers using an indirect encoding, which extends the representation of the chromosome to also include the NN topology, following a partial connectivity pattern. The authors show that the method was able to extract geometric information from the board that smoothed the learning of the network. 

Particularly relevant to this study are the applications of NE to general (video) game playing. In~\cite{reisinger2007coevolving}, Reisenger et al. successfully apply a co-evolutionary algorithm for NEAT in the General Game Playing competition~\cite{genesereth2005general} framework to evolve function approximation for game state evaluators. Hausknecht et al.~\cite{hausknecht2012hyperneat} introduced HyperNEAT-GGP to play two Atari games, Asterix and Freeway. The approach consisted in analyzing the raw game screen to detect objects that were used as input features for HyperNEAT, using the game score as fitness. This approach was later extended in~\cite{hausknecht2014neuroevolution} to a set of $61$ Atari games, comparing HyperNEAT and NEAT approaches in this learning task. Finally, Samothrakis et al.~\cite{samothrakis2015neuroevolution} used Separable Natural Evolution Strategies to evolve a NN that learns to play $10$ GVGAI games, showing that the methods proposed were able to learn most of the games played.

\section{NEAT} \label{sec:neat}

% NEAT involves two fundamental solutions to evolving artificial neural networks: \textit{innovation numbers} and \textit{speciation}.

%One is \textit{innovation numbers}, which allow for a more meaningful crossover. Another one is \textit{speciation}, so that generations are protected from being removed too early, Finally, NEAT starts the simplest network to incrementally making it more complex to find more efficient solutions~\cite{stanley2002evolving}. Each genome in NEAT has a list of NN nodes (input, hidden and output) and connections between them. Each connection gene contains an in-node, out-node, weight of connection, a boolean variable describing if its enabled or not, and an innovation number. 

% Figure~\ref{fig:neat} shows an abstract depiction of the NEAT genome encoding.
% \begin{figure}[!t]
% \centering 
%     \includegraphics[width=.49\textwidth]{Images/fig 2.png}
%     \caption{Schematic representation of the NEAT genome encoding~\cite{stanley2005real}.}
%     \label{fig:neat}
% \end{figure}

In NEAT, weights evolve following the established conventions in NeuroEvolution, using genetic operators as selection, crossover and mutation. The crossover operator generates offspring from two parents. In order to guarantee the appropriate crossing of individuals, NEAT uses \textit{innovation numbers}. These numbers are incremental values added to each connection and they act as indicators for when a gene appeared in the evolution process. Innovation numbers play a crucial role in the crossover operator as NEAT needs to be able to recombine genes with different topologies~\cite{radcliffe1993genetic}.

The offspring is formed either through uniform crossover, by randomly choosing matching genes, or through blended crossover, where the weights of two matching genes are averaged~\cite{wright1991genetic}. When two parents are selected to form an offspring, their genes are lined up based on their innovation numbers. Genes \textit{match} if they have the same innovation number. Genes that do not match are labelled as \textit{disjoint} (present in one parent but not the other) or \textit{excess} genes (those outside the range of a parent's innovation number). Dealing with these disjoint and excess genes is done by accepting genes directly from the more fit parent. In the case where both parents have equal fitness both their disjoint and excess genes are taken.

Mutations in NEAT can affect weights and the topology of the network in different ways, each one of them under certain probabilities:

\begin{itemize}
    \item Mutate link ($\mu_t$): creates a new link between two nodes.
    \item Mutate node ($\mu_n$): picks an existing random connection and splits it into two new connections with a new node in the middle; the original connection is disabled and the weight from the first node to the new middle node is set to $1.0$, while the connection from the middle to the second node is set to the original connection weight.
    \item Mutate weight shift probability ($\mu_{ws}$): alters the weight of a connection, shifted by a value picked uniformly at random factor in the range [$-W_s$, $W_s$].
    \item Mutate weight random probability ($\mu_{wr}$): replaces the weight of a connection by a value picked uniformly at random, in the range [$-W_r$, $W_r$].
    \item Mutate toggle link probability ($\mu_{tl}$): toggles a connection from enabled to disabled, and viceversa.
\end{itemize}

As can be seen, there are several types of mutations that can occur to the individuals. NEAT mutations, particularly the topological ones, form the basis for complexification of the evolved networks. Since smaller networks optimise faster and adding new mutations to genes can initially result in lower fitness scores, it can result in newer topologies having a small chance of surviving more than one generation. This can lead to losing innovation that could prove to be important in the future. In order to prevent this, NEAT uses \texttt{speciation}, which relies on the principle of populations within specific species competing against each other instead of competing against the entire population as a whole. This provides time for different topological innovations to optimise before they compete against other species. On each generation individuals are firstly organized into species according to a distance function. This function, shown in Equation~\ref{eq:species}, is a simple linear combination of the number of excess, disjoint genes and the average weight difference $\overline{W}$ of matching genes~\cite{stanley2002evolving}. 

\begin{equation}
    \delta=\frac{c_1 Excess}{N}+\frac{c_2 Disjoint}{N}+c_3 \cdot \overline{W}
    \label{eq:species}
\end{equation}

The coefficients $c_1$, $c_2$ and $c_3$ can be used to tune the impact of these 3 variables. $N$ is the number of connections from the larger genome, which is normalized to $1$ if there are less than $20$ connections. Each species is represented by a randomly chosen genome from that species in the previous generation. Individuals are compared to these representatives in the distance function. A new individual is placed in the first species with a distance as determined by Equation~\ref{eq:species}, respecting a maximum threshold $CP$. If the genome is not able to find a suitable species, a new species is created with this new individual as its representative.

% For a more detailed description of NEAT, the reader is referred to~\cite{stanley2002evolving}.

%Figure~\ref{fig:crossover} shows the genes of two equally fit parents lined up to crossover. 
% \begin{figure}[!t]
% \centering 
%     \includegraphics[width=.49\textwidth]{Images/fig 4.png}
%     \caption{Crossover of two equally fit parents forming an offspring  ~\cite{stanley2005real}.}
%     \label{fig:crossover}
% \end{figure}

\begin{figure} [!t]
\centering 
    \includegraphics[width=.49\textwidth]{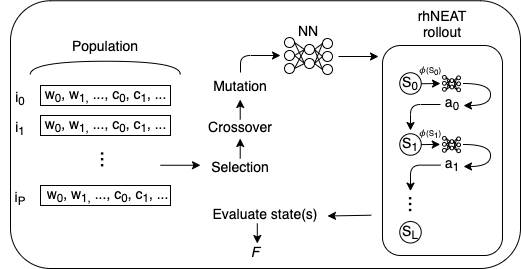}
    \caption{%This figure summarizes the evaluation of an individual. 
    Summary of an individual evaluation. 
    %After genetic operators are applied (tournament selection, uniform crossover) to the population, one individual is evaluated by taking the NN it represents and performing a rollout. Features $\phi(S_i)$ are extracted from each state $S_i$ and used as input for the NN, which returns an action to pass to the FM. This process repeats rolling the state forward $L$ times. The individual fitness $f$ is computed by evaluating one or more states visited during the rollout.
    }
    \label{fig:rollout}
\end{figure}

%%%%%%%%%%%%%%%%%%%%%%%%%%%%%%%%%%%%%%%
\section{Rolling Horizon NEAT for GVGAI} \label{sec:methods}

Rolling Horizon NEAT (rhNEAT) combines the statistical forward planning capabilities of RHEA with NEAT's ability to select actions based on game state features used as input for the NN. The approach is summarized as follows: rhNEAT evolves a population of individuals encoding nodes and links for a NN. The input of these NNs are game state features, and their output is a suggested action to apply in the given game state. Each NN is evaluated by rolling the game state forward for $L$ steps using the NN's output as actions. The value of one or more visited game states (see Section~\ref{sec:exp}) is assigned as fitness. Finally, rhNEAT returns the action suggested by the best individual (the NN with the highest fitness). Section~\ref{ssec:rhNEAT} explains rhNEAT in more detail and Section~\ref{ssec:inputs} describes the inputs and outputs used for this study.

\subsection{rhNEAT} \label{ssec:rhNEAT}

Rolling Horizon NEAT (rhNEAT) is configured to have a fixed population size of $P$ individuals. NEAT starts with the simplest network first to incrementally make it more complex through evolution. rhNEAT individuals are initialized with only a representation of the input and output nodes, and an empty connection map. As the NN evolves (selection by randomly choosing matching genes and uniform crossover), new connections are created, and their innovation number is updated. If using speciation, individuals are grouped into species as described in Section~\ref{sec:neat}. 

\begin{table}[!t]
\begin{center}
\caption{rhNEAT parameters and their values.}
% \resizebox{\columnwidth}{!}{%
\begin{tabular}{|c|c|c|}
\hline
\textbf{Parameter} & \textbf{Name} & \textbf{Value} \\
\hline\hline
$P$ & Population Size & $10$ \\
$L$ & Rollout length & $15$ \\
$R$ & Individuals discarded per generation & $20\%$ \\
$CP$ & Speciation Threshold & $4$ \\
$c_1$ & Excess coefficient & $1.0$ \\
$c_2$ & Disjoint coefficient & $1.0$ \\
$c_3$ & Weight difference coefficient & $1.0$ \\
$\mu_l$ & Mutate Link Probability & $0.5$ \\
$\mu_n$ & Mutate Node Probability & $0.3$ \\
$\mu_{ws}$ & Mutate Weight Shift Probability & $0.5$ \\
$W_s$ & Weight Shift Strength & $0.4$ \\
$\mu_{wr}$ & Mutate Weight Random Probability & $0.6$ \\
$W_r$ & Weight Random Strength & $1.0$ \\
$\mu_{tl}$ & Mutate Toggle Link Probability & $0.05$ \\
$FM_{b}$ & Forward Model calls budget & $1000$ \\
\hline 
\end{tabular}%
% }
\label{tab:params}
\end{center}
\end{table}

% \subsection{rhNEAT rollout}
One of the key characteristics of rhNEAT is the use of a Forward Model. An individual is evaluated by performing a rollout: the NN encoded is given features $\phi(S_i)$ extracted from the game state $S_i$ as input, and the action output $a_i$ is used to roll the game state forward; this process is repeated for $L$ steps (or until the game is over), see Figure~\ref{fig:rollout}.
Each visited state can be evaluated using a simple function, described in Equation~\ref{eq:heuristic}\footnote{This evaluation function is used by all rhNEAT configurations tested in the experiments of this paper, as well as by the RHEA and MCTS agents.}. The fitness $f$ of the individual can be computed by considering the evaluation (or reward) of the last state, or a combination of the rewards observed in the states visited, possibly discounted. Different configurations are experimented with in this paper to compute this evaluation (see Section~\ref{ssec:rewards}).

\begin{equation}\label{eq:heuristic}
h(s) = \begin{cases}
10^6 & \text{win = True}\\
-10^6 & \text{win = False}\\
\text{game score} & \text{otherwise}
\end{cases}
\end{equation}

The individuals in the population (or in each species, if using speciation) are sorted based on their fitness, and a percentage $R$ of the lowest scoring members are discarded at every generation. If speciation is used and a species has no individuals or only the representative left, said species is removed from the algorithm. 
%As in NEAT, rhNEAT uses genetic functions such as crossover and mutations. 

Once the computational budget is over, the algorithm selects the individual with the highest fitness and runs its network forward using features for the current game state as input. The output action is then returned to be played in the game. rhNEAT will then receive another call in the next frame to select an action and continue playing. At this stage, the population evolved in the previous game tick is initialized again to start a fresh evolutionary process. The algorithm can continue evolution from the previous population, an approach referred to as \texttt{population carrying} in this paper; the effects of keeping the population from one frame to the next are explored in Section~\ref{ssec:res_abl}. All tunable parameters of rhNEAT and their selected values are included in Table~\ref{tab:params}.

\subsection{Input and outputs} \label{ssec:inputs}

Unlike RHEA, rhNEAT requires game state inputs to evaluate an individual. The following game features are employed:

\begin{itemize}
    \item Avatar $x, y$ position, normalized in $[0,1]$.
    \item Avatar $x, y$ orientation.
    \item Avatar's health points, in $[0, M_{hp}]$, where $M_{hp}$ is the maximum health points achievable in each game.
    \item Proportion of up to three resources ${r_1, r_2, r_3}$ gathered by the avatar, where each $r_i$ is normalized in $[0,20]$.
    \item Distance $d$ and orientation $o$ to the closest instance of a sprite of the following categories:
    \begin{itemize}
        \item $d$ and $o$ to the closest NPC sprite.
        \item $d$ and $o$ to the closest Immovable sprite.
        \item $d$ and $o$ to the closest Movable sprite.
        \item $d$ and $o$ to the closest Resource sprite.
        \item $d$ and $o$ to the closest Portal sprite.
        \item $d$ and $o$ to the closest sprite produced by the avatar.
    \end{itemize}
    Distances are normalized in [0,$M_{d}$], where $M_{d}$ is the maximum possible distance in a game level. Orientation is normalized in $[-1,1]$, where $0$ represents that the distance vector to the sprite is aligned with the avatar's orientation and $-1$ when the sprite is in the opposite direction to the avatar's orientation. $o$ gradually progresses to $1$ with degrees to the right and to $-1$ to the left (i.e. $90$ degrees to the right corresponds to a value of $0 = 0.5$).
\end{itemize}

Health points, resources, distances and orientations to the different sprites are only considered if such features exist in the game (i.e. some games don't have NPCs, or health point systems), in order to reduce the input size of the network. However, in GVGAI games, it is possible that some observations do not appear before a certain frame (for instance, enemies that are spawned in Aliens are not visible in the first few frames). Carrying the population from one frame to the next is not straightforward when two consecutive frames have different input sizes. The approach taken in this implementation is to reinitialize the whole population when this (rarely) happens. The network output, however, is kept constant during the game, set to the number of actions available in the game\footnote{Some GVGAI games - not used in this study - may also change the number of available actions mid-game. For those cases, reinitializing the population would also be advisable.}.

\begin{table}[!t]
\begin{center}
\caption{Summary of results showing, for each approach, average win rate (and standard error) in the 20 games, the number of games it achieved the highest positive win rate in the subset (including the absolute highest, i.e. no ties, count), and the number of games in which it achieved the highest score.
%Summary of the results of the experiments presented in this paper. The first and second columns indicate the experimental sets and approaches as presented in Section~\ref{sec:exp}. The third column shows the average win rate across 20 games ($2000$ plays per approach) followed by the standard error of the measure. The fourth column shows the number of games each approach ranked as the one with highest positive win rate in the subset. Value within brackets shows the absolute highest (no ties) count. The last column shows the number of games each approach achieved the highest score.
}
% \resizebox{\columnwidth}{!}{%
% \begin{tabular}{|c|c|c|c|c|}
\begin{tabular}{|>{\centering\arraybackslash} m{0.15cm}|>{\centering\arraybackslash} m{2cm}|>{\centering\arraybackslash} m{1.85cm}|>{\centering\arraybackslash} m{1.7cm}|>{\centering\arraybackslash} m{1cm}|}
\hline
% & \textbf{Algorithm} & \textbf{Win Rate (Std Err)}  & \textbf{Highest positive win rates (absolute)} & \textbf{Highest score} \\
& \textbf{Algorithm} & \textbf{Win Rate\newline(std err)}  & \textbf{Highest $>0$ win rate (absolute)} & \textbf{Highest Score} \\
\hline\hline
\parbox[t]{2mm}{\multirow{4}{*}{\rotatebox[origin=c]{90}{Ablations}}} & baseline rhNEAT& $15.54\%$ $(6.13)$ & $0$ $(0)$ & $0$ \\
& rhNEAT(+sp) & $22.69\%$ $(7.01)$ & $2$ $(1)$ & $2$ \\
& rhNEAT(+cp) & $30.45\%$ $(7.93)$ & $2$ $(2)$ & $6$ \\
& rhNEAT(+sp,+cp) & $36.5\%$ $(8.52)$ & $13$ $(12)$ & $12$\\
\hline
\parbox[t]{2.5mm}{\multirow{3}{*}{\rotatebox[origin=c]{90}{Rewards}}} & rhNEAT & $36.5\%$ $(8.52)$ & $8$ $(6)$ & $12$\\
& rhNEAT-acc & $35.15\%$ $(8.12)$ & $9$ $(7)$ & $5$ \\
& rhNEAT-accdisc & $34.2\%$ $(8.14)$ & $3$ $(2)$ & $4$ \\
\hline
\parbox[t]{2mm}{\multirow{3}{*}{\rotatebox[origin=c]{90}{Fitness}}} & rhNEAT & $36.5\%$ $(8.52)$ & $10$ $(8)$ & $11$\\
& rhNEAT-lr & $35.25\%$ $(8.72)$ & $5$ $(4)$ & $3$ \\
& rhNEAT-avg & $31.55\%$ $(8.10)$ & $5$ $(2)$ & $6$ \\
\hline
\parbox[t]{2mm}{\multirow{4}{*}{\rotatebox[origin=c]{90}{All}}} & rhNEAT & $36.5\%$ $(8.52)$ & $3$ $(2)$ & $1$\\
& RHEA & $44.8\%$ $(8.89)$ & $2$ $(0)$ & $0$ \\
& MCTS & $42.65\%$ $(9.56)$ & $6$ $(3)$ & $4$ \\
\cdashline{2-5}
& SotA & $51.21\%$ $(8.72)$ & $13$ $(10)$ & $16$ \\
\hline
\end{tabular}
\label{tab:results}
\end{center}
\end{table}

\begin{figure*}[!t]
    \centering
    \includegraphics[width=.79\textwidth]{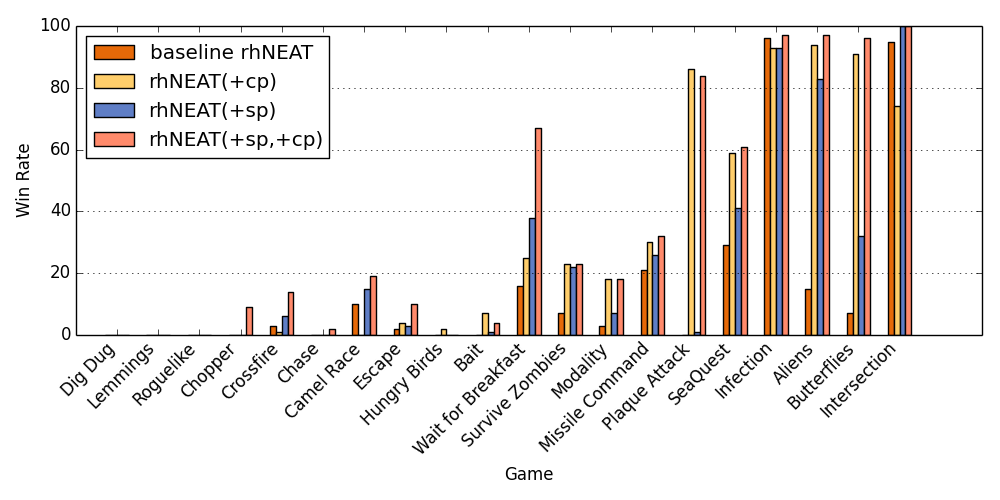}
    \caption{Win rates per game of different rhNEAT variants: basic rhNEAT, with population carrying (+cp), speciation (+sp) and with both (+cp,+sp).}\label{fig:abl}
\end{figure*}

%%%%%%%%%%%%%%%%%%%%%%%%%%%%%%%%%%%%%%%
\section{Experimental Setup} \label{sec:exp}

The objective of the experiments presented in this paper is to evaluate the performance of different variants of rhNEAT, as well as to ascertain how specific components of the algorithm contribute towards victory rate and average score. Experiments were run in 20 games of the GVGAI framework, the same ones used in previous studies~\cite{gaina2017enhancements}\cite{Gaina2018}, which represent an assorted selection of environments in terms of difficulty and game types. Each game is run $100$ times, testing $20$ repetitions on each of their $5$ levels.

A decision budget is given to rhNEAT that determines when evolution should stop before providing an action to be played in the game. This decision budget can take multiple forms, from a number of generations to actual wall time. The option chosen for this paper is to limit the number of usages of the Forward Model, so results are independent from the machine specifications where they are run and can also be compared with other algorithms. In order to provide a fair comparison with vanilla versions of other methods, rhNEAT is configured to run with a population size of $10$ individuals, a rollout length of $15$ and a budget of $1000$ Forward Model calls. 
%The list of specific rhNEAT parameters, hand tuned prior to the experiments described in this paper, can be seen in Table~\ref{tab:params}. 

Three sets of experimental studies have been conducted to explore rhNEAT variants: first, we explore the effect of different components in the algorithm, namely speciation and population carrying; second, we explore alternatives for calculating and assigning individual fitness, using the best variant from the first set of experiments as baseline; and third, we compare the best overall rhNEAT agent with other SFP methods and RHEA state of the art\footnote{\url{https://github.com/GAIGResearch/rhNEAT/} includes all code and results.}.

% \begin{figure*}[!t]
%     \centering
%     \includegraphics[width=.9\textwidth]{Images/rhneat-ablations.png}
%     \caption{Win rates per game of different rhNEAT variants: with population carrying and speciation (rhNEAT) and versions withouh speciation (-sp), population carrying (-cp) or both (-sc,cp).}\label{fig:abl}
% \end{figure*}

%%%%%%%%%%%%%%%%%%%%%%%%%%%%%%%%%%%%%%%
\section{Results and Discussion} \label{sec:results}

% Introduce section?

\subsection{rhNEAT Additions}\label{ssec:res_abl}

The objective of this set of experiments is to determine how incorporating certain components into rhNEAT (speciation and population carrying, as described in Section~\ref{sec:methods}) influences the performance of the algorithm. In order to differentiate the variants, we refer to \textit{baseline rhNEAT} as the version of the algorithm that does not use any of these components; \textit{rhNEAT(+cp)} refers to the version of the algorithm with population carrying; \textit{rhNEAT(+sp)} adds speciation; and \textit{rhNEAT(+sp,+cp)} incorporates both enhancements. All variants assign the value of the state found at the end of the rollout as the individual fitness.

The results clearly show that using speciation and population carrying increases both win rate and game scores. The first row group in table~\ref{tab:results}, while it's clear that \textit{baseline rhNEAT} performs poorly with only $15.54\%$ victory rate, it's interesting to observe the difference in performance when the two components are added. There is an increase in performance to $22.69\%$ observed when speciation is added, \textit{rhNEAT(+sp)}, while adding population carrying makes it further increase to $30.45\%$. Even more, the configuration with both features achieves a $36.5\%$ victory rate, with the highest number of best win rate across games ($13$, $12$ as the absolute best agent) and obtaining the best score in $12$ out of the $20$ games tested. These results suggest that the algorithm benefits from having different niches of weights in the population of rhNEAT, but even more when the complete population is kept between frames. Although this may not be surprising, it's important to highlight that, in GVGP, resetting the entire population \textit{may be} beneficial in some circumstances, in order to adapt to drastic changes in the game mechanics or the appearance/disappearance of features used as input for rhNEAT. 

Figure~\ref{fig:abl} shows the win rate for each one of the games and, in most cases, \textit{rhNEAT(+sp,+cp)} achieves the highest victory rate, with a similar ranking of the performance of the algorithms as observed in Table~\ref{tab:results}. It's interesting to highlight, like in most GVGAI studies, that the performance is very different across the distinct games, with some of them achieving close to $100\%$ win rate (\textit{Infection}, \textit{Aliens}, \textit{Butterflies}, \textit{Intersection}) and others close or equal to $0\%$ (\textit{Dig Dug}, \textit{Lemmings}, \textit{Roguelike}, \textit{Camel Race}, \textit{Crossfire}). A plausible explanation for this division is that the former games have a richer reward landscape, while the latter are either deceptive (in the case of \textit{Lemmings}), sparse (\textit{Camel Race}) or the games are particularly long (\textit{Dig Dug} and \textit{Roguelike}). However, as highlighted below, the little improvement in these hard games is one of the strengths of rhNEAT.

\begin{figure*}[!t]
    \centering
    \includegraphics[width=.79\textwidth]{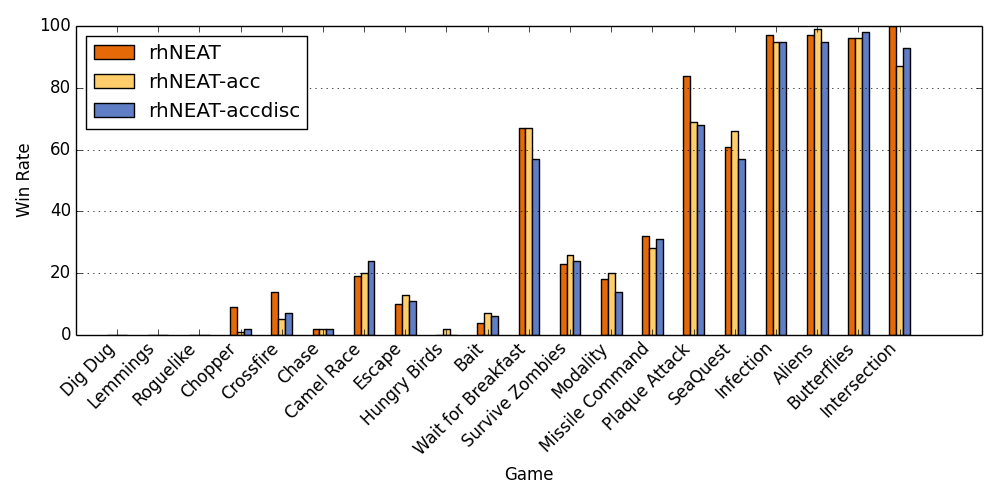}
    \caption{Win rates per game of different rhNEAT variants: using the evaluation of the last state of the rollout (rhNEAT), accumulated through all states visited (-acc) and accumulated and discounted (-accdisc).}\label{fig:heur}
\end{figure*}

\subsection{Reward and Fitness Alternatives}\label{ssec:rewards}

This section presents alternatives to how fitness is assigned to individuals. All variants use the best algorithm from the previous experiment (with speciation and population carrying). % introduce section combining these 2, all use best from part 1.

\subsubsection{rhNEAT Rewards} 

The first part of this experiment aims to analyze which procedure to compute the reward of a given rollout provides better results. Three different versions are compared: \textit{rhNEAT} uses the value of the game state at the end of the rollout as reward. \textit{rhNEAT-acc} and \textit{rhNEAT-accdisc} provide an accumulated sum of the values of all states visited during the rollout, with discount factor $\gamma=0.9$ for \textit{rhNEAT-accdisc}. This experiment is meant to compare short and long-term sight variants of rhNEAT, within the same rollout length.

There is not much difference between \textit{rhNEAT} and \textit{rhNEAT-acc}, as they both achieve a similar highest win rate counts. This suggests that rating individuals attending to either the evaluation of the state found at the end of the rollout or considering all the intermediate states does not impact performance significantly in terms of win rate. However, if the accumulated sum is discounted, the results show a drop in performance. Although the overall win rate is not significantly different to the other two options ($34.2\%$ versus $35.15\%$ for \textit{rhNEAT-acc} and $36.5\%$ for \textit{rhNEAT} - see second row group in Table~\ref{tab:results}), the number of games where \textit{rhNEAT-accdisc} achieves the highest win rate is clearly lower ($3$ versus $9$ and $8$ respectively). Figure~\ref{fig:heur} sheds some light into this discrepancy: per game, \textit{rhNEAT-accdisc} tends to achieve marginally worse results in most games, with \textit{rhNEAT} and \textit{rhNEAT-acc} normally achieving higher win rates.
Given that \textit{rhNEAT} obtains a higher count of games with the highest achieved scores, the version of the algorithm that only uses the final state's evaluation is considered to be better (and therefore used in the comparison against the alternative versions of rhNEAT and other approaches).

\begin{figure*}[!t]
    \centering
    \includegraphics[width=.79\textwidth]{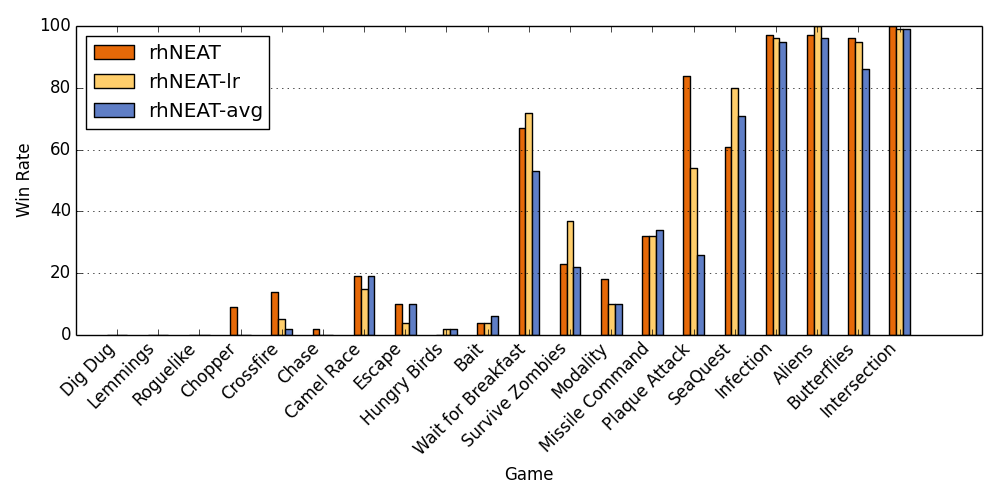}
    \caption{Win rates per game of different rhNEAT variants: individual fitness assigned as the last evaluation of the individual (rhNEAT), averaged for all evaluations (-avg) or updated with a learning rate (-lr).}\label{fig:fit}
\end{figure*}

\subsubsection{rhNEAT Fitness}

This experiment set focuses on the capacity to adapt to potential changes of the environment. While \textit{rhNEAT reward} provides a reward for a rollout, \textit{rhNEAT Fitness} explores how the individual fitness is computed form the reward. This reward can be assigned directly as the individual fitness (\textit{rhNEAT} variant) or, as individuals are likely to be evaluated multiple times during consecutive frames, averaged across. \textit{rhNEAT-avg} sets as individual fitness the arithmetic average of all the rewards seen by that individual. \textit{rhNEAT-lr} defines a learning rate ($\alpha=0.2$) so that, every time the individual receives a new fitness $f_i$, the individual fitness becomes $F = F + \alpha \times (f_i - F)$. While \textit{rhNEAT} focuses on the last experience only, the fitness of \textit{rhNEAT-avg} considers all past experiences and \textit{rhNEAT-lr} gives more weight to the more recent ones.

The results for different individual fitness are described in the third row group of Table~\ref{tab:results}. Computing the fitness as an exact average of the different rewards obtained by the individual (\textit{rhNEAT-avg}) seem to provide a lower win rate than the other two variants. This version is the one that gives less weight to the most recent evaluations, indicating that the algorithm benefits more from considering evaluations in the most recent game frames. One advantage is the potential adaptability of rhNEAT to changing situations in the environment. One disadvantage could be less robustness to stochastic environments. The average win rates are very similar between \textit{rhNEAT} and \textit{rhNEAT-lr}, but the former achieves the highest victory rate and score in more games than the latter, thus \textit{rhNEAT} is the configuration selected for the next comparison. 

Figure~\ref{fig:fit} shows the distribution of win rates per game for this setting. An interesting observation can be made about the games where there's an important difference in performance. \textit{rhNEAT-lr} clearly outperforms \textit{rhNEAT} in \textit{Seaquest} and \textit{Survive Zombies}, games where the density of points is high. A reasonable explanation for this could be that using a learning rate makes the algorithm favour those individuals that behave well with a rich reward landscape. \textit{rhNEAT} behaves better in games like \textit{Plaque Attack} and \textit{Crossfire}, where rewarding events (positive and negative) are more spread out across the game. Assigning the last rollout reward as the fitness of the individual seems to make rhNEAT more adaptable.

\subsection{Comparison with Other Algorithms} 

In order to analyze how rhNEAT compares to other methods used in the literature, the last set of experiments compares the results on the same games with those from sample GVGAI versions of \textit{RHEA} and \textit{MCTS}, using the same population, individual/rollout length, budget and state evaluation functions. Exploration constant for MCTS is set to $\sqrt{2}$. Additionally, \textit{rhNEAT} is compared with RHEA state-of-the-art (\textit{SotA}) results, as described in~\cite{gaina2020rolling}.

\begin{figure*}[!t]
    \centering
    \includegraphics[width=.79\textwidth]{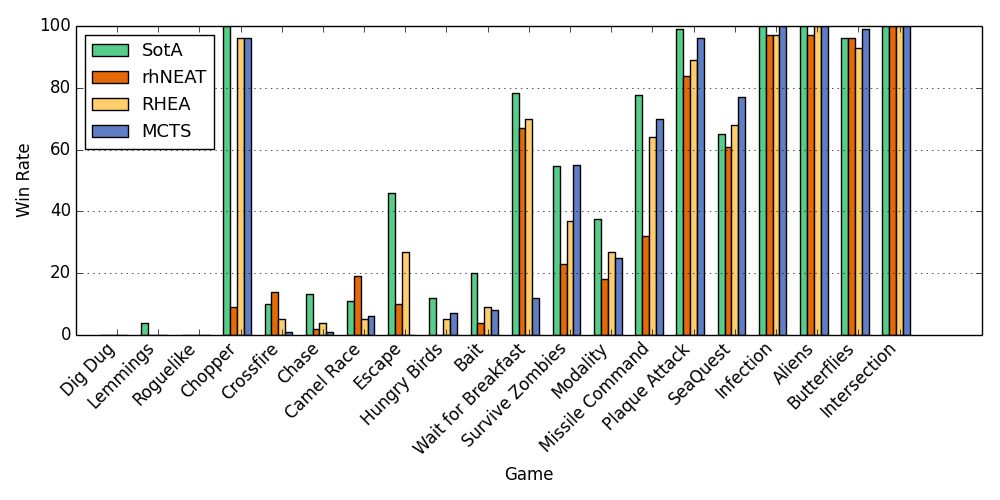}
    \caption{Win rates per game for rhNEAT, MCTS, RHEA and state of the art results for rolling horizon methods.}\label{fig:all}
\end{figure*}

Figure~\ref{fig:all} and the last rows of Table~\ref{tab:results} compare the best version of rhNEAT (with speciation, population carrying, and the last game state value used as individual fitness) with MCTS, RHEA and state-of-the-art results obtained by RHEA. In general, the performance of \textit{rhNEAT} is below that of the other comparable methods, which achieve $6$ (for \textit{MCTS}) and $8$ (for \textit{RHEA}) percentage points above \textit{rhNEAT}. \textit{SotA} achieves the highest standards, as it can be expected. However, it is important to note that \textit{SotA} aggregates results from multiple configurations of RHEA: while this comparison is illustrative, \textit{SotA} represents an upper bound not ever achieved by any single algorithm in particular. In any case, it is worthwhile highlighting that \textit{rhNEAT} achieves the highest positive win rate in three games (\textit{Intersection}, \textit{Crossfire} and \textit{Camel Race}), beating RHEA state of the art and MCTS in the last two of the games; these are games with sparse rewards, where game playing agents have normally struggled to play well~\cite{gaina2019tackling}. Results are also comparable in another $4$ games (right side of the plot in Figure~\ref{fig:all}) where all algorithms achieve high victory rates. \textit{rhNEAT} performs particularly worse than RHEA in Chopper and Escape, which can be attributed to the high number of sprites of the same type present in these games. A more curated feature selection could potentially improve results in this type of games.

% \subsection{Games}

%%%%%%%%%%%%%%%%%%%%%%%%%%%%%%%%%%%%%%%
% \section{Results and Discussion} \label{sec:results}

%Results for all experiments are summarized in Table~\ref{tab:results} and Figures~\ref{fig:abl} to~\ref{fig:all}. 
% Table~\ref{tab:results}
%shows the average win rate of all approaches across the $20$ games used for testing and a count on the number of games in which each algorithm achieves the higher victory rate (tied and uniquely). This table also includes the number of games where each algorithm obtains the highest average score across the $100$ repetitions per game. This table 
%provides an overall view of the performance of each method and their modifications. In contrast, Figures~\ref{fig:abl} to~\ref{fig:all} indicate the the win rate of each method per game.

%%%%%%%%%%%%%%%%%%%%%%%%%%%%%%%%%%%%%%%
\section{Conclusions} \label{sec:end}

This paper introduces rhNEAT, a new Statistical Forward Planning (SFP) algorithm that combines the concepts of Rolling Horizon Evolutionary Algorithms (RHEA) and of NeuroEvolution of Augmented Topologies (NEAT), and tests it in a variety of games from the General Video Game AI (GVGAI) corpus. The algorithm receives game features as input for a neural network (NN) that outputs one of the possible game actions. The architecture and weights of the NN are evolved with an evolutionary algorithm. Each individual (or NN configuration) is assigned a fitness by rolling the game forward, applying the actions dictated by the NN given the input features of each state, until the end of the \textit{rollout} is reached and the final game state is evaluated. The paper compares different configurations for rhNEAT and analyzes their performance across $20$ GVGAI games.

The best rhNEAT variant out of those explored in this paper uses speciation, population carrying, and assigns the value of the last game state reached in the rollout as the fitness of the individual. Results show that rhNEAT achieves a lower overall win rate than other SFP methods like Monte Carlo Tree Search and RHEA, although it is able to obtain better results than the state of the art in two games that are normally hard for this kind of methods. We believe that this result showcases the potential of this method, deserving more investigation in the future. One possibility for future work is to explore different and/or changing learning rates for \textit{rhNEAT-lr}, a variant that showed good performance. Another possible avenue is to dynamically alternate between different rhNEAT settings depending on the game state and the perceived reward landscape, as results showed that different variants seem to perform differently depending on these factors.

The literature review on NEAT variants is extensive~\cite{baldominos2019automated} and the present work only touches the surface of it. A further possibility could be to explore indirect representations such as Compositional Pattern Producing Networks (CCPNs) or HyperNEAT, as well as other methods for population control~\cite{baldominos2018evolutionary} or Convolutional layers to extract features from the game screen. Finally, the paradigm of combining general game features with forward planning can be extended to other methods, such as Grammatical Evolution~\cite{o2001grammatical}, Tangled Program Graphs~\cite{kelly2017multi} or different variants of Genetic Programming. 
%ADD note about opening the door to alternative representations?

%%%%%%%%%%%%%%%%%%%%%%%%%%%%%%%%%%%%%%%
\section*{Acknowledgments}
This work was funded by the EPSRC Centre for Doctoral Training in Intelligent Games and Game Intelligence (IGGI) EP/L015846/1. This research utilised Queen Mary's Apocrita HPC facility, supported by QMUL Research-IT. http://doi.org/10.5281/zenodo.438045.

\bibliographystyle{IEEEtran}
\bibliography{IEEEabrv,main}

\begin{thebibliography}{10}
\providecommand{\url}[1]{#1}
\csname url@samestyle\endcsname
\providecommand{\newblock}{\relax}
\providecommand{\bibinfo}[2]{#2}
\providecommand{\BIBentrySTDinterwordspacing}{\spaceskip=0pt\relax}
\providecommand{\BIBentryALTinterwordstretchfactor}{4}
\providecommand{\BIBentryALTinterwordspacing}{\spaceskip=\fontdimen2\font plus
\BIBentryALTinterwordstretchfactor\fontdimen3\font minus
  \fontdimen4\font\relax}
\providecommand{\BIBforeignlanguage}[2]{{%
\expandafter\ifx\csname l@#1\endcsname\relax
\typeout{** WARNING: IEEEtran.bst: No hyphenation pattern has been}%
\typeout{** loaded for the language `#1'. Using the pattern for}%
\typeout{** the default language instead.}%
\else
\language=\csname l@#1\endcsname
\fi
#2}}
\providecommand{\BIBdecl}{\relax}
\BIBdecl

\bibitem{genesereth2005general}
M.~Genesereth, N.~Love, and B.~Pell, ``General game playing: Overview of the
  aaai competition,'' \emph{AI magazine}, vol.~26, no.~2, pp. 62--62, 2005.

\bibitem{perez20152014}
D.~Perez-Liebana, S.~Samothrakis, J.~Togelius, T.~Schaul, S.~M. Lucas,
  A.~Cou{\"e}toux, J.~Lee, C.-U. Lim, and T.~Thompson, ``The 2014 general video
  game playing competition,'' \emph{IEEE Transactions on Computational
  Intelligence and AI in Games}, vol.~8, no.~3, pp. 229--243, 2015.

\bibitem{perez2019general}
D.~Perez-Liebana, J.~Liu, A.~Khalifa, R.~D. Gaina, J.~Togelius, and S.~M.
  Lucas, ``{General Video Game AI: A Multitrack Framework for Evaluating
  Agents, Games, and Content Generation Algorithms},'' \emph{IEEE Transactions
  on Games}, vol.~11, no.~3, pp. 195--214, 2019.

\bibitem{browne2014MCTSsurvey}
C.~Browne, E.~Powley, D.~Whitehouse, S.~Lucas, P.~Cowling, P.~Rohlfshagen,
  S.~Tavener, D.~Perez, S.~Samothrakis, and S.~Colton, ``{{A Survey of Monte
  Carlo Tree Search Methods}},'' in \emph{{IEEE Trans. on Computational
  Intelligence and AI in Games}}, vol.~4, no.~1, 2014, pp. 1--43.

\bibitem{perez2013rolling}
D.~Perez, S.~Samothrakis, S.~Lucas, and P.~Rohlfshagen, ``Rolling horizon
  evolution versus tree search for navigation in single-player real-time
  games,'' in \emph{Proceedings of the 15th annual conference on Genetic and
  evolutionary computation}, 2013, pp. 351--358.

\bibitem{perez2014knowledge}
D.~Perez, S.~Samothrakis, and S.~Lucas, ``Knowledge-based fast evolutionary
  mcts for general video game playing,'' in \emph{2014 IEEE Conference on
  Computational Intelligence and Games}, 2014, pp. 1--8.

\bibitem{joppen2017informed}
T.~Joppen, M.~U. Moneke, N.~Schr{\"o}der, C.~Wirth, and J.~F{\"u}rnkranz,
  ``Informed hybrid game tree search for general video game playing,''
  \emph{IEEE Transactions on Games}, vol.~10, no.~1, pp. 78--90, 2017.

\bibitem{gvgaibook2019}
D.~Perez-Liebana, S.~M. Lucas, R.~D. Gaina, J.~Togelius, A.~Khalifa, and
  J.~Liu, \emph{{General Video Game Artificial Intelligence}}.\hskip 1em plus
  0.5em minus 0.4em\relax Morgan \& Claypool Publishers, 2019, vol.~3, no.~2,
  \url{https://gaigresearch.github.io/gvgaibook/}.

\bibitem{gaina2017analysis}
R.~D. Gaina, J.~Liu, S.~M. Lucas, and D.~P{\'e}rez-Li{\'e}bana, ``Analysis of
  vanilla rolling horizon evolution parameters in general video game playing,''
  in \emph{European Conference on the Applications of Evolutionary
  Computation}.\hskip 1em plus 0.5em minus 0.4em\relax Springer, 2017, pp.
  418--434.

\bibitem{gaina2017enhancements}
R.~D. Gaina, S.~M. Lucas, and D.~P{\'e}rez-Li{\'e}bana, ``{Rolling Horizon
  Evolution Enhancements in General Video Game Playing},'' in \emph{IEEE
  Computational Intelligence and Games (CIG)}, 2017, pp. 88--95.

\bibitem{gaina2019tackling}
------, ``Tackling sparse rewards in real-time games with statistical forward
  planning methods,'' in \emph{Proceedings of the AAAI Conference on Artificial
  Intelligence}, vol.~33, 2019, pp. 1691--1698.

\bibitem{baldominos2019automated}
A.~Baldominos, Y.~Saez, and P.~Isasi, ``On the automated, evolutionary design
  of neural networks: past, present, and future,'' \emph{Neural Computing and
  Applications}, pp. 1--27, 2019.

\bibitem{stanley2002evolving}
K.~O. Stanley and R.~Miikkulainen, ``Evolving neural networks through
  augmenting topologies,'' \emph{Evolutionary computation}, vol.~10, no.~2, pp.
  99--127, 2002.

\bibitem{stanley2005real}
K.~O. Stanley, B.~D. Bryant, and R.~Miikkulainen, ``Real-time neuroevolution in
  the nero video game,'' \emph{IEEE transactions on evolutionary computation},
  vol.~9, no.~6, pp. 653--668, 2005.

\bibitem{gauci2010autonomous}
J.~Gauci and K.~O. Stanley, ``Autonomous evolution of topographic regularities
  in artificial neural networks,'' \emph{Neural computation}, vol.~22, no.~7,
  pp. 1860--1898, 2010.

\bibitem{reisinger2007coevolving}
J.~Reisinger, E.~Bahceci, I.~Karpov, and R.~Miikkulainen, ``Coevolving
  strategies for general game playing,'' in \emph{2007 IEEE Symposium on
  Computational Intelligence and Games}.\hskip 1em plus 0.5em minus 0.4em\relax
  IEEE, 2007, pp. 320--327.

\bibitem{hausknecht2012hyperneat}
M.~Hausknecht, P.~Khandelwal, R.~Miikkulainen, and P.~Stone, ``Hyperneat-ggp: A
  hyperneat-based atari general game player,'' in \emph{Proceedings of the 14th
  annual conference on Genetic and evolutionary computation}, 2012, pp.
  217--224.

\bibitem{hausknecht2014neuroevolution}
M.~Hausknecht, J.~Lehman, R.~Miikkulainen, and P.~Stone, ``A neuroevolution
  approach to general atari game playing,'' \emph{IEEE Trans. on CI and AI in
  Games}, vol. 6:4, pp. 355--366, 2014.

\bibitem{samothrakis2015neuroevolution}
S.~Samothrakis, D.~Perez-Liebana, S.~M. Lucas, and M.~Fasli, ``Neuroevolution
  for general video game playing,'' in \emph{2015 IEEE Conference on
  Computational Intelligence and Games (CIG)}, 2015, pp. 200--207.

\bibitem{radcliffe1993genetic}
N.~J. Radcliffe, ``Genetic set recombination and its application to neural
  network topology optimisation,'' \emph{Neural Computing \& Applications},
  vol.~1, no.~1, pp. 67--90, 1993.

\bibitem{wright1991genetic}
A.~H. Wright, ``Genetic algorithms for real parameter optimization,'' in
  \emph{Foundations of genetic algorithms}.\hskip 1em plus 0.5em minus
  0.4em\relax Elsevier, 1991, vol.~1, pp. 205--218.

\bibitem{Gaina2018}
R.~D. Gaina, S.~M. Lucas, and D.~Perez-Liebana, ``{General Win Prediction from
  Agent Experience},'' in \emph{Computational Intelligence and Games (CIG),
  2018 IEEE Conference on}, 2018, p. to appear.

\bibitem{gaina2020rolling}
R.~D. Gaina, S.~Devlin, S.~M. Lucas, and D.~Perez-Liebana, ``{Rolling Horizon
  Evolutionary Algorithms for General Video Game Playing},''
  \emph{arXiv:2003.12331}, 2020.

\bibitem{baldominos2018evolutionary}
A.~Baldominos, Y.~Saez, and P.~Isasi, ``Evolutionary convolutional neural
  networks: An application to handwriting recognition,'' \emph{Neurocomputing},
  vol. 283, pp. 38--52, 2018.

\bibitem{o2001grammatical}
M.~O'Neill and C.~Ryan, ``Grammatical evolution,'' \emph{IEEE Transactions on
  Evolutionary Computation}, vol.~5, no.~4, pp. 349--358, 2001.

\bibitem{kelly2017multi}
S.~Kelly and M.~I. Heywood, ``Multi-task learning in atari video games with
  emergent tangled program graphs,'' in \emph{Proceedings of the Genetic and
  Evolutionary Computation Conference}, 2017, pp. 195--202.

\end{thebibliography}

\end{document}